\documentclass[runningheads]{llncs}
\usepackage[T1]{fontenc}
\usepackage[utf8]{inputenc}
\usepackage{graphicx,verbatim}
\usepackage{booktabs}
\usepackage{multirow}

\usepackage{amsmath}
\usepackage{amssymb}
\usepackage{booktabs}
\usepackage{multirow}

\usepackage[colorlinks]{hyperref}
\usepackage{color}
\usepackage{xcolor}
\hypersetup{colorlinks,
allcolors=black}

\begin{document}

\title{Re-mixing Embeddings for Patient Augmentation in Data Scarce Multiple Instance Learning}
\titlerunning{RECIPE}

\newcommand{\cofirst}{\textsuperscript{*}}
\newcommand{\cocorr}{\textsuperscript{\#}}

\author{Muhammed Furkan Dasdelen\inst{1}\cofirst \and
Fatih Ozlugedik\inst{1}\cofirst \and
Anastasia Litinetskaya\inst{1} \and
Nassir Navab\inst{2,3} \and
Carsten Marr\inst{1,3,4,5,6}\cocorr \and
Ario Sadafi\inst{1,2,3}\cocorr}
\authorrunning{Dasdelen et al.}
%
\institute{Computational Health Center \& Helmholtz AI, Helmholtz Munich, Munich, Germany \and
Computer Aided Medical Procedures, Technical University of Munich, Munich, Germany \and
Munich Center for Machine Learning (MCML), Munich, Germany \and
Department of Medicine III, Ludwig Maximilian University Hospital, Munich, Germany \and
Department of Physics, Ludwig-Maximilians-Universität München, Munich, Germany \and
DKTK, German Cancer Consortium, partner site Munich, Germany
}

\maketitle              

\begingroup
\renewcommand{\thefootnote}{}
\footnotetext{
\textsuperscript{*}Equal contribution \\
\textsuperscript{\#}Co-corresponding: \email{\{carsten.marr,ario.sadafi\}@helmholtz-munich.de}
}
\endgroup

\begin{abstract}

Data scarcity is a major bottleneck in medical Multiple Instance Learning (MIL), especially for rare diseases or expensive modalities. We introduce a statistically grounded patient augmentation approach that generates realistic patients directly in embedding space. Using Gaussian Mixture Models as a probabilistic clustering approach on pooled instance embeddings from all patients, our method learns disease-specific "recipes"—statistical distributions of instances across unsupervised clusters. New patients are then generated by sampling embeddings from clusters based on learned recipes. Unlike existing methods that require examples from all categories, our method can generate patients offline by re-mixing pooled embeddings. Generated patients are further selected based on uncertainty quantification to improve MIL performance. We evaluate our method across three clinically relevant scarcity scenarios: (i) cross-dataset transfer, where an entirely missing "healthy" class is generated using statistics from an external cohort; (ii) low-data regimes, where class sizes are extremely limited; and (iii) small-cohort non-image tasks, including single-cell RNA-seq and flow cytometry. Across all experiments, our method improves performance over baseline, often outperforming other bag-mixing strategies. Notably, in the missing-class scenario, a performance comparable to full-dataset training is achieved, demonstrating its potential for rare disease diagnostic and privacy-preserving patient augmentation. The code is available at \url{https://github.com/marrlab/RECIPE}



\end{abstract}
\section{Introduction}
    Multiple Instance Learning (MIL) has emerged as a powerful paradigm for weakly supervised learning in medical imaging, where only bag-level labels (e.g., patient diagnoses) are available and instance-level annotations (e.g., individual cell or image patch labels) are impractical to obtain \cite{Song2023}. In MIL, a “bag” comprises a set of instances—such as histopathology patches~\cite{lu2021data}, single cell images~\cite{hehr2023explainable}, or genomic/immune features ~\cite{litinetskaya2024multimodal,ding2026application}—and the learning objective is to predict the bag label while weighening key instances.
    
    One persistent bottleneck in healthcare MIL is data scarcity, especially in rare diseases and in specialized, expensive technologies or settings that require invasive data collection. Traditional augmentation techniques—rotations, flips, and color perturbations—operate at the instance level and fail to generate realistic variation at the bag level. We introduce Re-mixing Embeddings via Clustering for Patient augmEntation (RECIPE), which generates new bags for MIL pipelines in data-scarce settings. Given disease-level labels from internal or external datasets, RECIPE unsupervisedly extracts instance “recipes” needed to form each disease class and enables uncertainty quantification. We showcase our pipeline in three scenarios: (i) when one class has no patients and the recipe is derived from an external dataset, (ii) when only a few patients exist in one class, and (iii) when the dataset is naturally small, as in single-cell RNA-seq or flow cytometry.
    
\section{Related work}
   
    Data augmentation improves generalization in machine learning~\cite{DBLP:journals/corr/ZhangBHRV16,Vapni2000TheNO,MUMUNI2022100258survey,bestpracticecnn}, yet its application in MIL is limited. Since whole slide images (WSI) contain largely non-discriminative patches, naive instance-level augmentation often amplifies noise and redundancy. Moreover, in MIL, patch-level augmentation entails significant computational overhead due to the need for feature re-extraction.
    
    To address this, bag-level strategies have been developed to generate new bags by mixing instances \cite{psemix,yang2022remixgeneralefficientframework}. PseMix generates synthetic bags in a mini-batch by combining patches from two WSIs into one mixed bag; the label is mixed in the same proportion. ReMix similarly augments within the training batch, but first summarizes each WSI into a few representative patch groups and then mixes those summaries. While these methods increase diversity, they face critical limitations. First, they operate exclusively online, producing ephemeral synthetic bags that cannot be validated or reused. Second, their local mixing scope fails to capture global population statistics. Most critically, they require real examples from every class, rendering them ineffective when an entire class is missing—a frequent challenge when a healthy control is missing and in rare disease research.
    
    We introduce RECIPE, a paradigm for offline, statistically-grounded patient generation. Unlike local mixing, RECIPE models global instance distributions via clustering to learn class-specific "recipes" of disease phenotypes. This enables the creation of persistent, analyzable cohorts. Uniquely, RECIPE supports cross-dataset transfer, generating missing classes using external statistics without sharing sensitive data. To our knowledge, this is the first MIL framework addressing missing patient classes, with applicability extending to single-cell and cytometry data.

\begin{figure}[t]
\includegraphics[width=\textwidth]{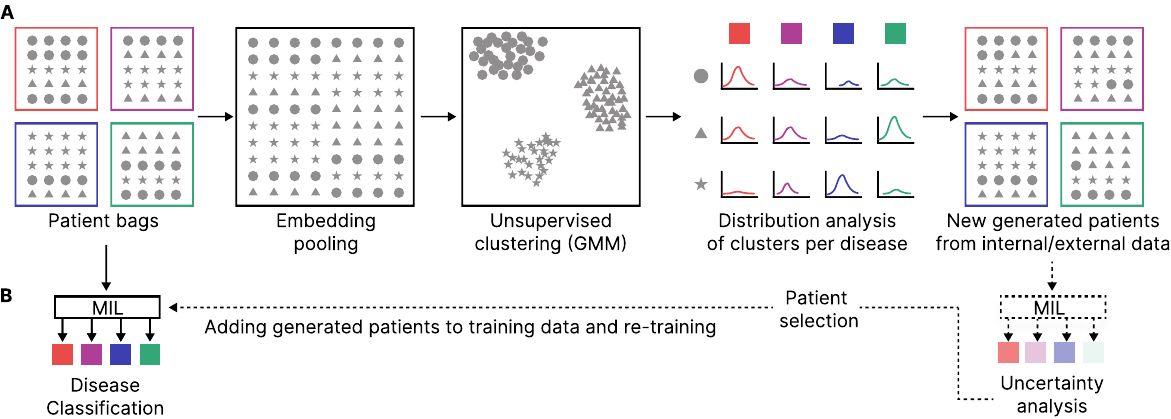}
\caption{\textbf{Re-mixing embeddings via clustering for patient augmentation (RECIPE)}. \textbf{(A)} A Gaussian Mixture Model (GMM) is fitted to pooled instance embeddings to define unsupervised clusters. Class-specific \textbf{"recipes"} are computed as distributions across these clusters (mean $\pm$ s.d.). New patients are generated by sampling instances according to these recipes. \textbf{(B)} Informative patients are selected by quantifying patient-level uncertainty via Monte Carlo dropout. The model is then retrained using the most uncertain generated samples to improve performance.} \label{fig1}
\end{figure}

\section{Methods}
    
    \subsection{Dataset and Preprocessing}
    We evaluated RECIPE on both image (hematologic cytology) and non-image (single-cell RNA-seq, flow cytometry) datasets where limited sample sizes benefit from patient-level augmentation.
    
    \textbf{AML-Hehr}~\cite{hehr2023explainable} contains single white blood cell (WBC) images from 189 patients, including healthy donors ($n=60$) and four AML genetic subtypes ($n=129$; \textit{CBFB::MYH11}, \textit{NPM1}, \textit{PML::RARA}, \textit{RUNX1::RUNX1T1}), collected at Munich Leukemia Laboratory (MLL) (2009–2020). 
    
    \textbf{cAItomorph}~\cite{dasdelen2026ai} comprises peripheral blood smear images from 2043 patients across seven hematological conditions and healthy donors, collected at MLL (2021–2022). Images were resized to $224 \times 224$, normalized (ImageNet statistics), and embedded into 768-dimensional vectors using the DinoBloom-B~\cite{koch2024dinobloom} hematology foundation model.
    
    \textbf{PBMC}~\cite{stephenson2021single} is a single-cell RNA-seq cohort of 130 donors (647,366 cells) with varying COVID-19 severity. Following~\cite{litinetskaya2024multimodal}, we utilized three major classes: healthy ($n=23$), mild ($n=23$), and severe ($n=13$). Cell embeddings were generated using PCA (30-dim) and scVI \cite{lopez2018deep} (50-dim; default parameters with \textit{site} and \textit{sample ID} covariates).
    
    \textbf{Covid-flow}~\cite{liechti2022immune} consists of flow cytometry data from COVID-19 patients and healthy controls. Merging mild and moderate classes~\cite{ding2026application} resulted in 172 healthy, 43 mild/moderate, and 54 severe cases. We used BDC1 and BDC2 panels with channel-wise z-score normalization.
    
    All datasets were evaluated using 5-fold cross-validation.
    
\subsection{Cluster-based unsupervised sampling}

        Our pipeline has two parts. In part~1 (Fig.~\ref{fig1}A), we pool instance embeddings, $\{\mathbf{x}_i\}_{i=1}^N \subset \mathbb{R}^D$, from all training patients, and fit a $K$-component Gaussian Mixture Model (GMM) using \textit{scikit-learn (v1.5)}:
        \begin{equation}
        p(\mathbf{x}_i)=\sum_{k=1}^K \pi_k\,\mathcal{N}\!\left(\mathbf{x}_i \mid \boldsymbol{\mu}_k,\boldsymbol{\Sigma}_k\right),
        \end{equation}
        where $\sum_{k=1}^K \pi_k=1$ and $\boldsymbol{\Sigma}_k$ is diagonal. We maximize the data log-likelihood via Expectation--Maximization (EM).

        For each real patient $p$ with class $c$, we count the instances assigned to each cluster $k$. Aggregating these yields the class--cluster mean $\mu_{c,k}$ and standard deviation $\sigma_{c,k}$, defining a class-specific \emph{recipe}. To generate a patient of class $c$, we sample $n_{c,k} \sim \mathcal{N}(\mu_{c,k},\sigma_{c,k}^2)$ and set $n_{c,k} \leftarrow \max(0,n_{c,k})$, then draw $n_{c,k}$ embeddings from the pool of real class-$c$ instances assigned to cluster $k$. The sampled embeddings form a new bag, generating new patients by re-mixing instances according to disease-specific statistics.
    
    \subsection{Uncertainty aware patient selection}
    
    Part~2 (Fig.~\ref{fig1}B) of the pipeline selects the most informative generated patients to improve MIL classification performance and is independent of Part~1. After generating patients, we estimate predictive uncertainty of the pretrained MIL model on these bags using Monte Carlo (MC) dropout at inference, with dropout applied in the classifier head. For $t=1,\ldots,T$ stochastic forward passes, we compute
    \begin{equation}
    \mathbf{z}^{(t)}(\mathbf{x}) = f(\mathbf{x};\,\boldsymbol{\theta}^{(t)}) \in \mathbb{R}^{C},
    \qquad
    \mathbf{p}^{(t)}(\mathbf{x}) = \mathrm{softmax}\!\left(\mathbf{z}^{(t)}(\mathbf{x})\right),
    \end{equation}
    and form the mean predictive distribution. We quantify uncertainty $U$ with the predictive entropy of $\bar{\mathbf{p}}(\mathbf{x})$:
        \begin{equation}
        \bar{\mathbf{p}}(\mathbf{x}) \;=\; \frac{1}{T}\sum_{t=1}^{T}\mathbf{p}^{(t)}(\mathbf{x}),
        \qquad
        U(\mathbf{x})
        \;=\;
        H\!\left[\bar{\mathbf{p}}(\mathbf{x})\right]
        \;=\;
        -\sum_{c=1}^{C}\bar{p}_{c}(\mathbf{x})\,\log \bar{p}_{c}(\mathbf{x}).
        \end{equation}
    where higher entropy indicates more uncertain predictions \cite{shannon1948mathematical,gal2017deep}. We also ablate BALD (Bayesian Active Learning by Disagreement) \cite{houlsby2011bayesian} and the standard deviation of the predicted class's maximum logit (Max-STD) across posterior samples \cite{gal2017deep} as alternative uncertainty metrics (see \ref{ablation}).

\begin{figure}[t]
\includegraphics[width=\textwidth]{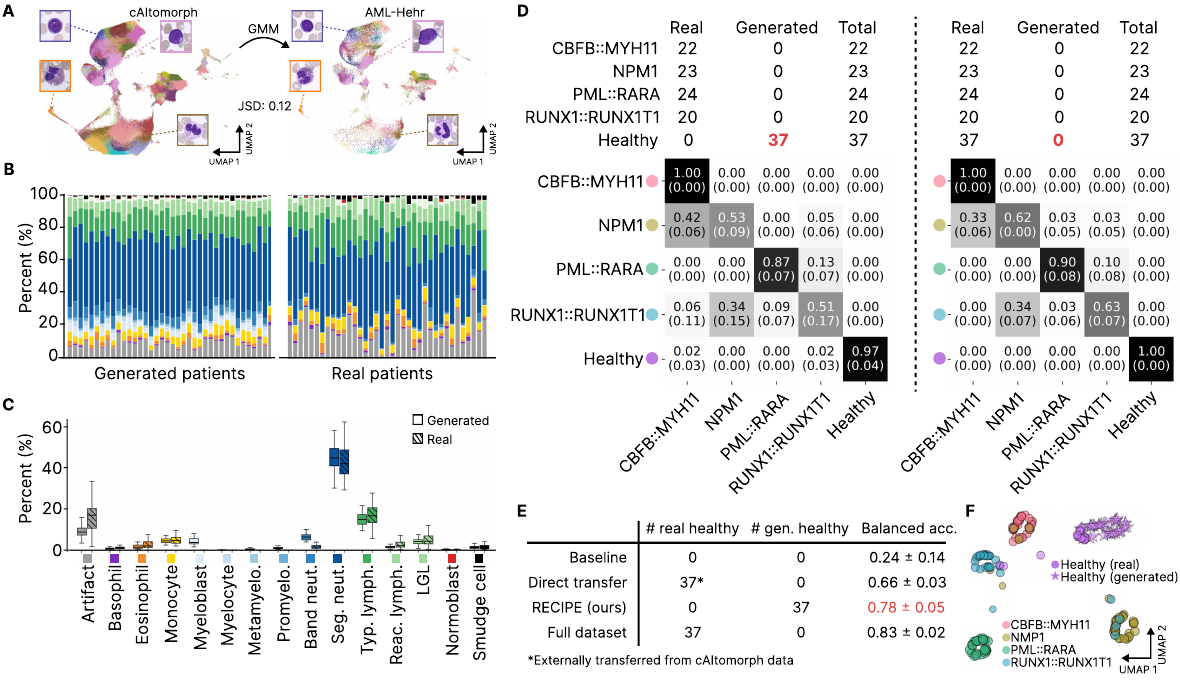}
\caption{\textbf{Generating realistic patients using externally derived recipes.} \textbf{(A)} We fit a Gaussian Mixture Model (GMM) on pooled instance embeddings from the cAItomorph data to obtain unsupervised clusters and class-specific \emph{recipes}. We then transfer this clustering structure to AML-Hehr and generate healthy individuals by sampling only from the AML-Hehr embedding pool according to the externally learned healthy recipe. Transferring the GMM yields aligned clusters across datasets: real cell-label distributions within corresponding clusters agree well between cAItomorph and AML-Hehr (JSD$=0.12$). \textbf{(B)} Per-patient cell-type composition of generated versus real AML-Hehr patients, computed using real cytomorphology labels. \textbf{(C)} Cell-type proportions across patients shows that generated patients recapitulate the real cell-composition statistics. \textbf{(D, E)} Training with generated patients achieves test-set performance comparable to the full-dataset setting. \textbf{(F)} Generated patients align well with real patients in the latent space, as visualized by UMAP.} \label{fig2}
\end{figure}

\section{Experiments}
    Previous patient-level augmentation approaches have mainly targeted already large cohorts (e.g., TCGA), where gains are often marginal \cite{psemix,yang2022remixgeneralefficientframework}. In practice, medical datasets are frequently imbalanced, rare classes and healthy controls can be difficult to collect, and some diagnostic tests are invasive or too costly to apply broadly. We therefore evaluate our pipeline in three clinically relevant data-scarce scenarios:
    
    \textit{(i) Generating patient recipes from an external dataset.} We intentionally removed all healthy controls from the AML-Hehr training split and derived the healthy recipe from healthy controls in the cAItomorph dataset. Using this external recipe, we then generated healthy individuals by sampling only from the AML-Hehr diseased instance embedding pool. We generated the same number of healthy controls as in the original AML-Hehr dataset ($n=37$). We could not include baseline augmentation comparisons in this setting because other existing methods assume that at least some real samples are available for every class.
    
    \textit{(ii) Generating patients from few samples.} We reduced the number of healthy controls in the AML-Hehr training set to $n \in {1,2,4,8,16,32}$ and generated additional healthy patients to match the original class size. In this setting, the recipe statistics were estimated from the few available healthy samples, while new bags were created by sampling instances from the full AML-Hehr embedding pool. We compared our method with other augmentation methods. 
    
    \textit{(iii) Application in single cell omics/cytometry.} We applied our approach to modalities where sample sizes are naturally limited, including single-cell RNA-seq and flow cytometry. In these experiments, we augmented the training set with generated patients corresponding to 30\% of the original training data size (see \ref{ablation}) and compared with other augmentation methods.
    
    In all experiments, number of clusters in GMM is fixed to $K=50$ (see \ref{ablation}).

\subsection{Multiple instance learning architecture and training}

Since our pipeline is architecture-agnostic, we used attention-based multiple instance learning (ABMIL)~\cite{ilse2018attention}. We also show compatibility with other aggregators such as Transformer \cite{wagner2023transformer} and DSMIL \cite{li2021dual} (see \ref{ablation}). ABMIL uses a 256-dimensional hidden representation with an MLP classifier. We trained ABMIL with batch size 16 and learning rate $1\times10^{-5}$ using AdamW (weight decay = 0.01) for up to 150 epochs, with early stopping based on validation loss. Training used a single NVIDIA A100 80GB GPU.

\section{Results}

\begin{figure}[t]
\includegraphics[width=\textwidth]{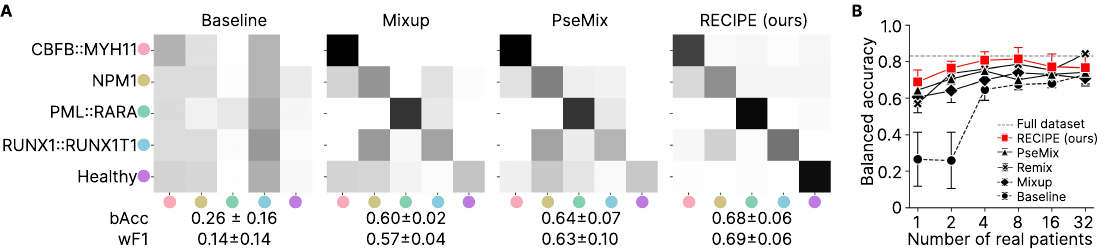}
\caption{\textbf{Augmenting classes with few patients.} \textbf{(A)} Performance comparison of augmentation methods when only one real healthy control is available in the training set. \textbf{(B)} RECIPE achieves the highest performance for almost all sample sizes.} \label{fig3}
\end{figure}

\subsection{Generating patient recipes from an external dataset}
    We fit a GMM on cAItomorph and computed healthy-control statistics (\emph{recipes}) from the resulting clusters, without using cell-type labels. We then applied the same GMM to AML-Hehr to assign clusters and generated healthy individuals using the precomputed recipe (Fig.~\ref{fig2}A). The clustering structure transferred well: corresponding clusters in the two datasets showed similar real cell-type compositions, with a low Jensen--Shannon divergence (0.12). Using the original AML-Hehr cell labels for evaluation, generated individuals closely matched the expected healthy blood composition (Fig.~\ref{fig2}B,C), with segmented neutrophils comprising $\sim$60\% and typical lymphocytes $\sim$20\%. We generated 37 healthy controls (matching the original cohort size) and trained ABMIL using these cases. On a test set of real patients, training with generated healthy controls achieved performance close to the full dataset ($0.78 \pm 0.05$ vs.\ $0.83 \pm 0.02$) and outperformed directly transferring healthy cases from cAItomorph to AML-Hehr ($0.66 \pm 0.03$).

\subsection{Generating patients from few samples}
    Next, we evaluated our pipeline in a low-data setting where only a few healthy controls are available. Disease recipes were estimated from the limited healthy samples ($n \in \{1,2,4,8,16,32\}$), and healthy individuals were generated from the pooled AML-Hehr embedding set based on these recipes. For a single healthy individual ($n=1$), the baseline achieved only $0.26 \pm 0.16$ balanced accuracy (Fig. \ref{fig3}A). MixUp and PseMix substantially improved performance ($0.60 \pm 0.02$ and $0.64 \pm 0.07$, respectively), while RECIPE achieved the best result ($0.68 \pm 0.06$). RECIPE also improved sensitivity across classes, particularly for the healthy class (Fig. \ref{fig3}A). This trend was consistent across all sample sizes (Fig. \ref{fig3}B).

\begin{table}[t]
\centering
\caption{Balanced accuracy (bAcc) results on non-image COVID-19 datasets. Values are mean $\pm$ std. Best result in each column is shown in bold.}
\label{tab:covid_nonimage_bacc}
\setlength{\tabcolsep}{8pt}
\scalebox{0.95}
{
\begin{tabular}{lcccc}
\toprule
& \multicolumn{2}{c}{\textbf{scRNA-seq}} & \multicolumn{2}{c}{\textbf{Flow cytometry}} \\
\cmidrule(lr){2-3}\cmidrule(lr){4-5}
\textbf{Method} & \textbf{PCA} & \textbf{scVI} & \textbf{BDC1} & \textbf{BDC2} \\
\midrule
Baseline
& $0.56 \pm 0.12$
& $0.58 \pm 0.15$
& $0.50 \pm 0.05$
& $0.49 \pm 0.05$ \\

Mixup
& $0.60 \pm 0.14$
& $0.65 \pm 0.12$
& $0.57 \pm 0.04$
& $0.58 \pm 0.04$ \\

ReMix
& $0.54 \pm 0.08$
& $0.58 \pm 0.10$
& $0.56 \pm 0.05$
& $0.55 \pm 0.05 $ \\

PseMix
& $0.63 \pm 0.14$
& $0.62 \pm 0.04$
& $0.57 \pm 0.05$
& $0.56 \pm 0.04$ \\

\textbf{RECIPE (ours)}
& $\mathbf{0.64 \pm 0.05}$
& $\mathbf{0.73 \pm 0.09}$
& $\mathbf{0.61} \pm 0.07$
& $\mathbf{0.60} \pm 0.03$ \\
\bottomrule
\end{tabular}
}
\end{table}

\begin{table}[t]
    \centering
    \caption{One-factor-at-a-time ablation study on Covid-19 scVI embeddings (left) and RECIPE performance with various aggregators on Covid-19 PCA embeddings (right). The first row denotes default settings; each subsequent row varies a single hyperparameter. Best balanced accuracy is shown in bold. Default: predictive entropy uncertainty, global most uncertain selection, 30\% augmentation ratio, and $K=50$ clusters}
    \begin{minipage}[t]{0.48\textwidth}
        \centering

        \label{tab:ablation_oat}
        {
        \setlength{\tabcolsep}{8pt} 
        \begin{tabular}{lc}
        \toprule
        \textbf{Hyperparameter} & \textbf{bAcc} \\
        \midrule
        Default & $\mathbf{0.73 \pm 0.09}$ \\
        \midrule
        Uncertainty: BALD & $0.62 \pm 0.23$ \\
        Uncertainty: Max-STD & $0.61 \pm 0.24$ \\
        \midrule
        mixed most uncert. & $0.72 \pm 0.17$ \\
        per-class most uncert. & $0.68 \pm 0.19$ \\
        random certain & $0.70 \pm 0.16$ \\
        most certain & $0.53 \pm 0.12$ \\
        \midrule
        Aug. \%: 5 & $0.68 \pm 0.16$ \\
        Aug. \%: 10 & $0.69 \pm 0.20$ \\
        Aug. \%: 20 & $0.70 \pm 0.15$ \\
        Aug. \%: 60 & $0.68 \pm 0.21$ \\
        Aug. \%: 90 & $0.70 \pm 0.14$ \\
        \midrule
        $K$: 3   & $0.61 \pm 0.16$ \\
        $K$: 10  & $0.60 \pm 0.10$ \\
        $K$: 30  & $0.52 \pm 0.16$ \\
        $K$: 100 & $0.48 \pm 0.17$ \\
        \bottomrule
        \end{tabular}
        }
    \end{minipage}%
    \hfill 
    \begin{minipage}[t]{0.50\textwidth}
        \centering
        \vspace{0pt} 
        \setlength{\tabcolsep}{8pt} 
        \begin{tabular}{lc}
        \toprule
        \textbf{Architecture} & \textbf{bAcc} \\
        \midrule
        ABMIL (Baseline)        & $0.56 \pm 0.12$ \\
        ABMIL (RECIPE)       & $0.64 \pm 0.05$ \\ 
        Improvement (\%) & $+15\%$ \\ 
        \midrule
        DSMIL (Baseline)        & $0.45 \pm 0.11$ \\
        DSMIL (RECIPE)         & $0.58 \pm 0.17$ \\
        Improvement (\%) & $+29\%$ \\ 
        \midrule
        Transformer (Baseline) & $0.49 \pm 0.09$ \\
        Transformer (RECIPE) & $0.67 \pm 0.11$ \\
        Improvement (\%) & $+38\%$ \\
        \bottomrule
        \end{tabular}
    \end{minipage}
\end{table}

\subsection{Application in single cell omics/cytometry}
    Finally, we evaluated RECIPE on real-world non-image modalities using scRNA-seq and flow cytometry (Table~\ref{tab:covid_nonimage_bacc}). On scRNA-seq, RECIPE achieved the best performance for both embedding types, improving balanced accuracy from $0.56 \pm 0.12$ to $0.64 \pm 0.05$ with PCA ($+14\%$) and from $0.58 \pm 0.15$ to $0.73 \pm 0.09$ with scVI ($+26\%$). On flow cytometry, RECIPE improved performance on both panels ($+22\%$): for BDC1, balanced accuracy increased from $0.50 \pm 0.05$ to $0.61 \pm 0.07$, for BDC2, RECIPE increased balanced accuracy from $0.49 \pm 0.05$ to $0.60 \pm 0.03$.

\subsection{Ablation studies}\label{ablation}

    We conducted a one-factor-at-a-time ablation study (Table \ref{tab:ablation_oat}) to evaluate the impact of key hyperparameters. Predictive entropy proved to be the most effective uncertainty measure, outperforming BALD and Max-STD by over 10\%. We also examined the selection scope, which determines how uncertain patients are prioritized: \textit{global} (top uncertain patients across the entire dataset), \textit{per-class} (top uncertain patients within each class), or \textit{mixed}. Global selection achieved the highest bAcc ($0.73 \pm 0.09$), suggesting that prioritizing the most uncertain samples regardless of class labels provides the most information. To demonstrate the benefit of selecting the most uncertain patients, we also ablated augmenting with the most certain patients or randomly selected patients. Selecting the most certain patients yielded the lowest performance, while selecting random patients resulted in intermediate performance. The augmentation ratio showed an optimal peak at 30\%; exceeding this threshold did not improve more. For the number of clusters, $K=50$ provided the best resolution for capturing semantic diversity. Finally, RECIPE is architecture-agnostic, providing significant performance gains—up to 38\%—across different aggregators including ABMIL, DSMIL, and Transformer.

\section{Conclusion}
    We propose RECIPE, a patient-level augmentation framework designed to overcome data scarcity in MIL. By deriving probabilistic class "recipes" via unsupervised clustering, our method generates realistic patients by re-mixing instance embeddings. A key innovation of RECIPE is its ability to generate entirely missing cohorts using externally derived statistics, demonstrating that disease phenotypes can be effectively transferred across datasets without sharing patient data. Furthermore, RECIPE proved highly effective in non-imaging domains, yielding significant performance gains in data-limited single-cell RNA-seq and flow cytometry tasks.

\begin{credits}
\subsubsection{\ackname} C.M. received funding from the European Research Council under the European Union's Horizon 2020 Research and Innovation Programme (grant agreements 866411, 101113551, and 101213822) and support from the Hightech Agenda Bayern.

\subsubsection{\discintname} The authors have no competing interest.

\subsubsection*{Author contributions} Conceptualization: AS, CM, MFD, FO; Data curation: AL, MFD; Methodology and software: FO, MFD; Writing-original draft: MFD, FO; Writing–editing: AS, NN, CM; Supervision: AS, NN, CM.

\end{credits}

\bibliographystyle{splncs04}
\bibliography{references}

\end{document}